\documentclass[10pt,twocolumn,letterpaper]{article}
\usepackage[dvipsnames,svgnames,x11names]{xcolor}
\definecolor{citecolor}{RGB}{119,185,0} 
\usepackage[pagebackref=false,breaklinks=true,letterpaper=true,colorlinks,citecolor=citecolor,bookmarks=false]{hyperref}
\usepackage{iccv}  
\usepackage{cite}
\usepackage{times}
\usepackage{epsfig}
\usepackage{graphicx}
\usepackage{amsmath}
\usepackage{amssymb}
\usepackage{algorithm}
\usepackage{algorithmic}
\usepackage{comment}
\usepackage{array}
\usepackage{url}
\usepackage{animate}
\usepackage{colortbl}
\usepackage{subfiles}

\usepackage{soul}
\usepackage[utf8]{inputenc}
\usepackage[small]{caption}
\usepackage{multirow}

\usepackage{pifont}
\def\eg{\emph{e.g.}} 
\def\ie{\emph{i.e.}} 
\def\etal{\emph{et~al.}} 

\newcommand{\resblock}[2]{\multirow{3}{*}{\(\left[\begin{array}{c}\text{3$\times$3, #1}\\[-.1em] \text{3$\times$3, #1} \end{array}\right]\)$\times$#2}
}

\newcommand{\convblock}[2]{\multirow{1}{*}{\([\begin{array}{c}\text{#2$\times$#2, #1} \end{array}]\)}
}

\newlength\savewidth\newcommand\shline{\noalign{\global\savewidth\arrayrulewidth
  \global\arrayrulewidth 1pt}\hline\noalign{\global\arrayrulewidth\savewidth}}
  
\iccvfinalcopy 
\begin{document}
\title{3D Magic Mirror: Clothing Reconstruction from a Single Image\\ via a Causal Perspective}
\author{Zhedong Zheng$^1$ \quad Jiayin Zhu$^1$ \quad Wei Ji$^1$ \quad Yi Yang$^2$ \quad Tat-Seng Chua$^1$ \\
$^1$Sea-NExT Joint Lab, National University of Singapore  \quad $^2$ Zhejiang University }
\maketitle

\begin{abstract}
This research aims to study a self-supervised 3D clothing reconstruction method, which recovers the geometry shape and texture of human clothing from a single image. Compared with existing methods, we observe that three primary challenges remain:
(1) 3D ground-truth meshes of clothing are usually inaccessible due to annotation difficulties and time costs; 
(2) Conventional template-based methods are limited to modeling non-rigid objects, \eg,  handbags and dresses, which are common in fashion images;
(3) 
The inherent ambiguity compromises the model training, such as the dilemma between a large shape with a remote camera or a small shape with a close camera.  

In an attempt to address the above limitations, we propose a causality-aware self-supervised learning method to adaptively reconstruct 3D non-rigid objects from 2D images without 3D annotations. In particular, to solve the inherent ambiguity among four implicit variables, \ie, camera position, shape, texture, and illumination, 
we introduce an explainable structural causal map (SCM) to build our model. The proposed model structure follows the spirit of the causal map, which explicitly considers the prior template in the camera estimation and shape prediction. When optimization, the causality intervention tool, \ie, two expectation-maximization loops, is deeply embedded in our algorithm to (1) disentangle four encoders and (2) facilitate the prior template. Extensive experiments on two 2D fashion benchmarks (ATR and Market-HQ) show that the proposed method could yield high-fidelity 3D reconstruction. Furthermore, we also verify the scalability of the proposed method on a fine-grained bird dataset, \ie, CUB. The code is available at \url{https://github.com/layumi/3D-Magic-Mirror}.
\vspace{-.1in}
\end{abstract}


\section{Introduction}
Nowadays, 
people can purchase clothing via online shopping sites, \eg,  Amazon and eBay. 
However, there remains a gap between the display images and the real product quality~\cite{liu2012street,chen2020image}. In an attempt to minimize such a visualization gap, we study the 3D clothing reconstruction from a single image. Given a 2D clothing image and the foreground mask, we intend to reconstruct a 3D mesh, which recovers the geometry shape and texture of the target clothing. Besides, the clothing reconstruction can also be applied to many computer vision applications, including virtual reality ~\cite{sing2016garden}, interactive system~\cite{ren2020interactive, sha2020progressive} and 3D printing~\cite{chae2015emerging}.  

\begin{figure}[tbp]
\begin{center}\vspace{-.25in}
    \includegraphics[width=0.95\linewidth]{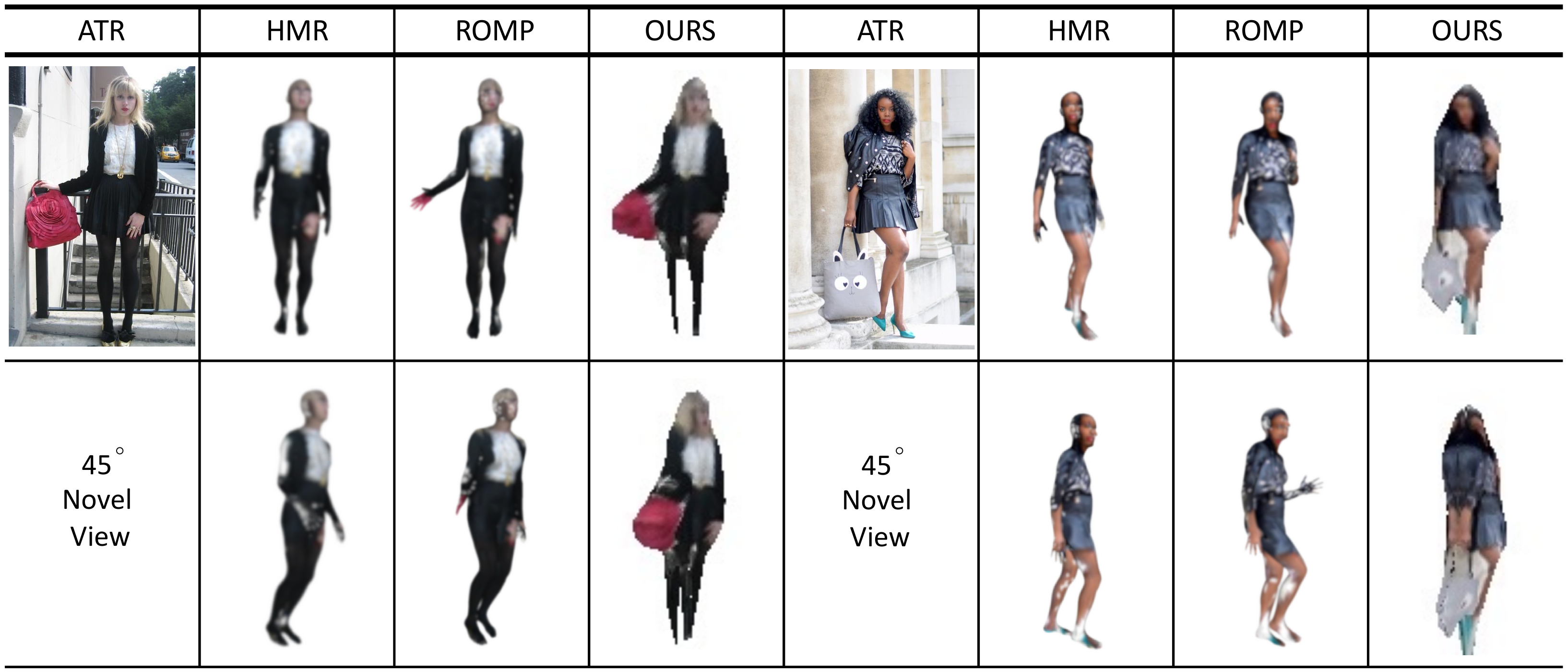}
\end{center}
\vspace{-.2in}
     \caption{ Motivation. Here we compare the proposed approach with prevailing template-based methods, \ie, HMR~\cite{kanazawaHMR18} and ROMP~\cite{ROMP} on a fashion dataset ATR~\cite{ATR}. We re-implement and visualize results with the color mapping according to the projected location. The first row is the front view, and the second row is the 3D mesh rotated with 45$^{\circ}$. The template-based model can capture the human poses but miss non-rigid objectives, such as hairs, handbags, and dresses. 
     }\label{fig:motivation}
\vspace{-.15in}
\end{figure}

\begin{figure}[tbp]
\begin{center}\vspace{-.25in}
    \includegraphics[width=1\linewidth]{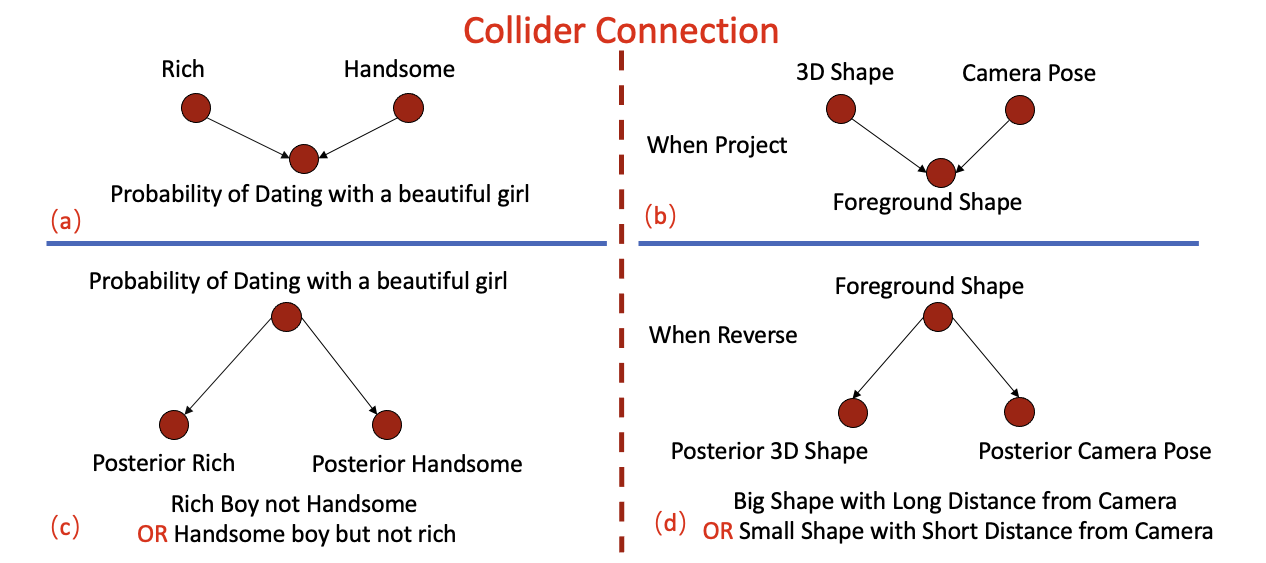}
\end{center}
\vspace{-.2in}
     \caption{ Explanation of the Collider Connection. Here we show a common example ``Dating with Beautiful Girl''~\cite{pearl2018book} in (a), which is similar to our simplified dilemma on two variants, \ie, shape and camera pose in (b). Since we study an inverse problem, we also draw (c) and (d). We note that there is a compensation effect when we estimate the posterior probability. As shown in (d), given the observed foreground shape, the model needs to estimate the posterior 3D shape and posterior camera pose. There are two possible alternatives for the network to learn, \ie, a big shape from a long distance or a small shape from a short distance. Therefore, it makes the model difficult to converge an answer.     
     }\label{fig:collider}
\vspace{-.15in}
\end{figure}

However, there remain three challenges. \textbf{First,} 3D annotations are difficult to obtain due to the annotation difficulty and time costs. There are no public large-scale 3D clothing mesh datasets for supervised learning. In contrast, the availability of the large-scale 2D fashion datasets, such as ATR~\cite{ATR} and Market-HQ~\cite{zheng2015scalable}, makes training data-hungry deep-learned approaches become feasible. The success has been proved in the 2D pedestrian image generation~\cite{ma2017pose, qian2018pose, ge2018fd,sarkar2021humangan,fu2022stylegan}.  
With the recent development in self-supervised learning and deep-learned models, one straightforward idea is raised whether we can leverage 2D data for 3D reconstruction, even without manual 3D annotations. 
\textbf{Second,} existing works~\cite{kanazawaHMR18,lin2021end,lin2021mesh} typically focus on human pose estimation and body reconstruction via parametric models, \eg,  a morphable body template~\cite{SMPL:2015}. However, pre-defined body parameters usually are not scalable to non-rigid clothing, \eg, dresses, handbags, and loose clothing~\cite{xiu2022icon,EVA3D}, losing fine-grained clothing details. As shown in Figure~\ref{fig:motivation}, we re-implement two prevailing methods, \ie, HMR~\cite{kanazawaHMR18} and ROMP~\cite{ROMP}, which both successfully capture the human pose but miss the cloth shape. In contrast, our methods leverage a deformable model to further facilitate learning non-rigid objects.  \textbf{Third,} one scientific question still remains in single image 3D reconstruction. The primary implicit variables for reconstruction are camera viewpoint, shape, texture, and illumination. Ideally, these four factors are independent. However, it remains challenging to disentangle the implicit variables in practice. 
One typical dilemma is the ambiguity between the camera and shape~\cite{li2020umr}. Given a 2D image, it is hard to decide the object size. There are two possible answers (see Figure~\ref{fig:collider}). One large object is far from the camera or one small object is close to the camera. Despite different physical sizes, the two objects have the same projection size in photos.       


\begin{figure*}[t]
\begin{center}\vspace{-.3in}
    \includegraphics[width=1\linewidth]{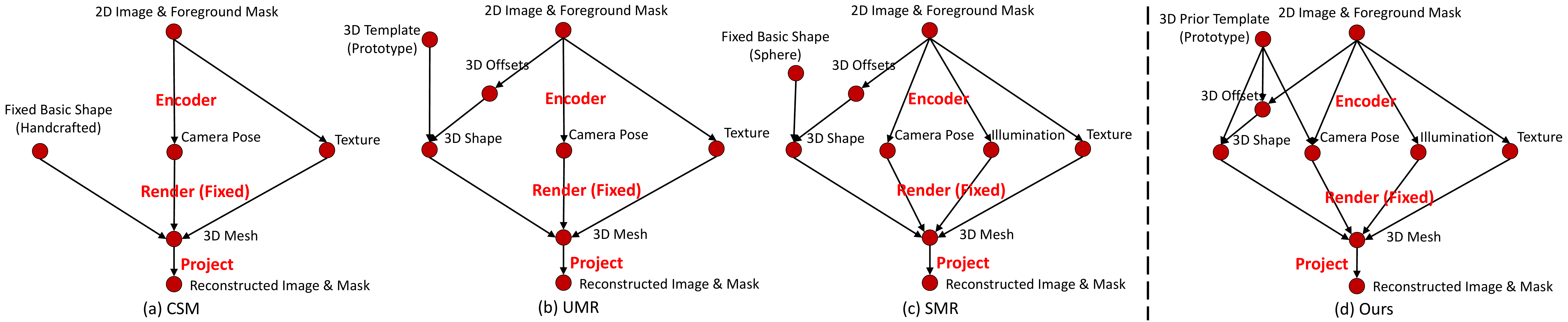}
\end{center}
\vspace{-.25in}
     \caption{The Structural Causal Map (SCM). We compare the proposed method with three typical 3D reconstruction works, including CSM~\cite{kulkarni2019canonical}, UMR~\cite{li2020umr}, and SMR~\cite{hu2021smr}. Here we show the image reconstruction loop, \ie, ``2D $\rightarrow$ 3D $\rightarrow$ 2D''. (a) Given one 2D image and the foreground mask, CSM only applies two encoders for Camera and Texture, since it assumes that the basic shape is shared, ignoring intra-class changes. (b, c) UMR and SMR further introduce the shape encoder, which is to predict the shape offset. The final shape is obtained by adding the shape offset to the shape template. (d) In this work, we argue that the predicted shape offsets are also conditioned on the prior template. Besides, the prior template also impacts the camera prediction. Therefore, we explicitly introduce the dependency with the prior prototype. It is worth noting that our prototype is initialized from an ellipsoid and iteratively updated during training, which does not require any hand-crafted initialization. (Qualitative comparison with more other methods is listed in Table~\ref{table:methods}.) 
     }\label{fig:causal}
\vspace{-.2in}
\end{figure*}

In an attempt to overcome the above-mentioned challenges, this paper proposes a self-supervised learning approach to adaptively reconstruct 3D clothing from a single image without 3D mesh annotation. We study the existing works (see Figure~\ref{fig:causal}) and introduce an explainable structural causal map (SCM) to build our model and guide the optimization strategy.  
(1) Following the spirit of the causal map, we deploy four independent 3D attribute encoders and a differentiable render for reconstruction. The encoders are to extract 3D attributes from 2D images and foreground masks, including camera viewpoint, geometric shape, texture, and illumination. Then the attributes are fed to the differentiable render to reconstruct the 3D mesh. Different from existing works~\cite{hu2021smr, li2020umr}, we explicitly introduce the prior template to help the camera estimation and shape offsets estimation, which is aligned with human observation. If we foreknow the human  prototype, it helps us to predict the camera position as well as the intra-class variant (such as leg movement).  
(2) We leverage the causality intervention tool, \ie, two expectation-maximization loops, to help learn the prototype, and disentangle encoders from the confusing loss punishment. 
To summarize, our contributions are two-fold: 
\begin{itemize}
\vspace{-2.5mm}
    \item We identify the three challenging problems in the 3D clothing reconstruction: 1) No 3D annotations; 2) Non-rigid objects; 3) Reconstruction ambiguity. In an attempt to solve these challenges, we propose a self-supervised learning method with a causality design to reconstruct the 3D clothing mesh from large-scale 2D image datasets. Following the spirit of the causal map, we re-design the encoder structure and leverage the ``intervention'' tool, \ie, two expectation-maximization loops, to facilitate the 3D attribute encoder learning.
\vspace{-2.5mm}
    \item Experiments on two fashion datasets verify the effectiveness of the proposed method quantitatively and qualitatively. Furthermore, experiments on the fine-grained bird dataset also show that the proposed method has good scalability to other non-rigid objects.  
\vspace{-2.5mm}
\end{itemize}

\setlength{\tabcolsep}{3pt}
\begin{table}\vspace{-.25in}
\caption{Comparison with existing methods on supervisions. The proposed method harnesses relatively weak supervision for 3D reconstruction from a single image.  It is also worth noting that some works take 2D input images with white background as inputs, and we also view this line of works deploying the foreground mask. }
\label{table:methods}\vspace{-.25in}
\begin{center}\resizebox{\linewidth}{!}{\footnotesize
\begin{tabular}{l|c|c|c|c|c}
\hline
\multirow{2}{*}{Methods} & Viewpoint   & Semantic & Manual   & Part  & Foreground  \\
& Annotations & Keypoint & Template & Seg.  & Mask        \\
\shline
 VPL~\cite{kato2019learning}                    & \checkmark &            &            &          &\checkmark\\
 CMR~\cite{kanazawa2018learning}$^\dagger$ $^*$ &(\checkmark)& \checkmark &(\checkmark)&          &\checkmark\\
 CSM~\cite{kulkarni2019canonical}               &            &            & \checkmark &          &\checkmark\\
 DIB-R~\cite{chen2019dibr}             & \checkmark &            &            &          &\checkmark\\
 IMR~\cite{tulsiani2020implicit}$^*$            &            &            &(\checkmark)&          &\checkmark\\
 ACMR-vid~\cite{li2020online} &            &     \checkmark &            & &\checkmark\\
 UMR~\cite{li2020umr}                           &            &            &            &\checkmark&\checkmark\\
 WLDO~\cite{biggs2020wldo}                      &            & \checkmark & \checkmark &          &\checkmark\\
 Texformer~\cite{xu20213d}                      &            &            & \checkmark &\checkmark&\checkmark\\
MeshInversion~\cite{zhang2022monocular}                        &            &            &            &          &\checkmark\\
 SMR~\cite{hu2021smr}                           &            &            &            &          &\checkmark\\
 \hline
 Ours & & & & & \checkmark \\
\hline
\end{tabular}}
\end{center}\vspace{-.15in}
\footnotesize{$^*$: The method deploys the manual template for initialization; \\ $^\dagger$: The viewpoint annotation is optional. 
\vspace{-.15in}
}
\end{table}

\section{Related Work}\label{sec:relatedwork}
\begin{figure}[tbp]
\begin{center}\vspace{-.25in}
    \includegraphics[width=1\linewidth]{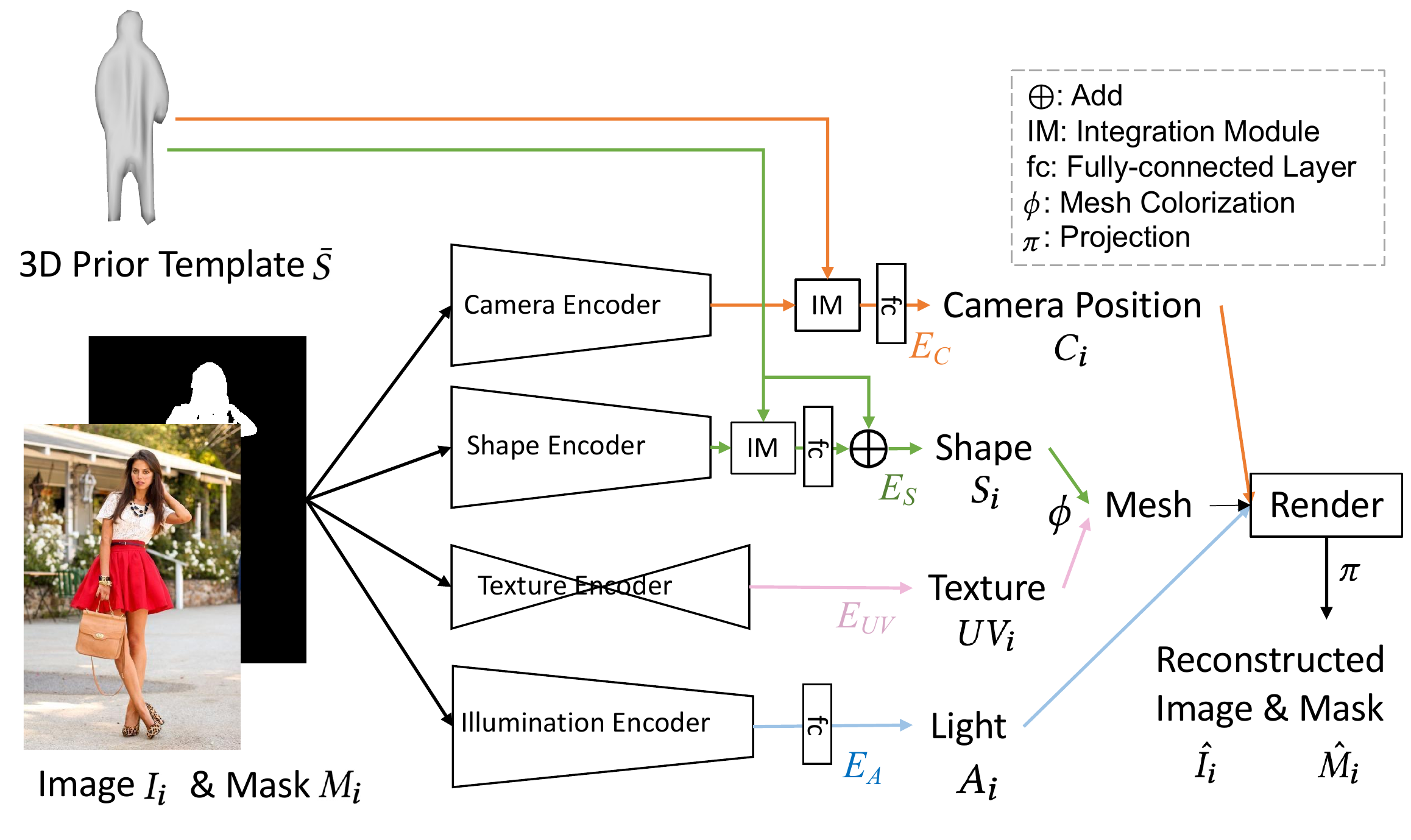}
\end{center}
\vspace{-.3in}
     \caption{ Overview. Here we show a ``2D$\rightarrow$3D$\rightarrow$2D'' loop. We follow the causal map in Figure~\ref{fig:causal} (d) to design the pipeline. Given one pair of clothing image $I_i$ and the mask $M_i$, we deploy four independent encoders $E_c$, $E_s$, $E_{UV}$, and $E_A$ for camera position, shape, texture and light estimation. We introduce the integration module (IM) to fuse the local feature from 3D prior template $\bar{S}$. Then we apply the colorize function $\phi$ to obtain the mesh, and utilize the render to re-project the mesh into 2D space via the project function $\pi$. Finally, we obtain the reconstructed $\hat{I_i}$ and $\hat{M_i}$. During inference, we manipulate intermediate camera attributes $C_i$ to generate novel-view images of the target person.
     }\label{fig:pipeline}
\vspace{-.2in}
\end{figure}

\noindent\textbf{3D Reconstruction from Single Image.}
Humans can estimate 3D structures from a single image. Many works deploy a parameter-based template~\cite{kanazawaHMR18,ROMP,lin2021mesh}, which is robust but also limits the representative ability to non-rigid objects. To enable more degrees of freedom, Deephuman~\cite{zheng2019deephuman} adopts a U-Net model to reconstruct the human body and clothing voxel, but dense depths and ground-truth 3D annotations are needed. To reduce the dependency on 3D annotations, PrGANs~\cite{gadelha20173d} trains a generative adversarial network (GAN) to generate the 3D voxel from lots of 2D images with different viewpoints. Tulsiani \etal~\cite{tulsiani2017multi} further leverage multi-view photos of the identical object to reconstruct the 3D voxel model in an unsupervised manner. To avoid the dense prediction of the voxel format, Pixel2Mesh~\cite{pixel2mesh,wen2019pixel2mesh++} is a fully supervised learning work with the well-designed regularization, which reconstructs mesh by deforming an ellipsoid.  
VPL~\cite{kato2019learning} leverages viewpoint annotation to ensure mesh reconstruction consistency via adversarial training. 
To invade the viewpoint annotation, one of the early works is Canonical Surface Mapping (CSM) ~\cite{kulkarni2019canonical}, which treats 3D reconstruction as a dense key-point estimation problem. 
However, CSM deploys a fixed pre-defined shape template, which largely limits intra-class shape changes for different instances. To address the shape limitation, CMR~\cite{kanazawa2018learning} first proposes to use a learnable shape template and Li~\etal~\cite{li2020umr} further proposes UMR, which adopts a two-stage training strategy for template updating. 
The segmentation parsing~\cite{hung2019scops} is used in UMR for better alignment. The contemporary work, IMR~\cite{tulsiani2020implicit}, 
first introduces the mapping function instead of the vertex location regression, saving computation costs. 
Taking a further step, SMR~\cite{hu2021smr} aligns the 3D mesh via mix-up and conducting the auxiliary vertex classification, while MeshInversion~\cite{zhang2022monocular} deploys foreground prediction. The proposed method is mainly different from existing works in three aspects: (1) \textbf{Weaker supervision.} As shown in Table~\ref{table:methods}, the proposed method demands limited supervision and mainly leverages the large-scale multi-view images to learn prior knowledge.  
(2) \textbf{Model design.} As shown in Figure~\ref{fig:pipeline}, the network design follows the causal map. 
(3) \textbf{Optimization strategy.} To deal with the compensation effect in the loss punishment, we deploy the ``intervention'' tool, \ie, two expectation-maximization loops, to facilitate the 3D attribute learning.

\noindent\textbf{Causal Learning.} 
Causal learning is to identify causalities from a set of empirical factors, which can be either pure observations or counterfactual inference~\cite{peters2017elements}. 
To represent causalities, a causal model is usually defined via structural equations and graphs~\cite{pearl2009causality}. 
According to the structural causal model, manipulations can be conducted to optimize the estimated relations between variables, \eg, ``Do'' operation is to cut certain directed edges and control the target variable~\cite{pearl2009causality}. One line of works using the counterfactual thought is to conduct data augmentation~\cite{chen2020counterfactual,mao2021generative,wei2022synthesizing,li2022invariant,ren2022dice} and obtain the debiased prediction ~\cite{niu2021counterfactual,zhang2020causal,tian2022debiasing}. 
Another line of works on the generative model mainly explores implicit causal learning to discover the causal factors during training~\cite{zhang2021causerec,liu2022structural}. CausalGAN~\cite{kocaoglu2017causalgan} trains a generator, which is consistent with an implicit causal graph, and is able to sample from either conditional labels or interventional distributions. Similarly, CausalVAE~\cite{yang2021causalvae} is a VAE-based causal framework, which discovers latent causal factors in data with graph constraints. 
Both methods implicitly harness causal mapping by learning latent code or adding one graph constraint. However, the causality learned from data is not always accurate and explainable, limiting the causality effect. 
Differently, 
our model follows the spirit of causality between semantic entities to (1) explicitly consider the causality relation between entities, \eg, leveraging the prior template to help both camera encoder and shape encoder learning; and (2) explicitly apply  ``intervention'' tools to solve the ambiguity of learning multiple variables. 


\noindent\textbf{Expectation Maximization (EM).} 
EM is an iterative method to find the parameters with maximum likelihood in statistical models ~\cite{moon1996expectation}. The EM algorithm iteratively conducts two kinds of steps: an expectation step (E-step) to obtain the expectation of latent variables and a maximization step (M-step) which computes parameters to maximize the expected likelihood based on the latent variables. Since the M-step updates parameters, it affects the E-step in the next round. In this way, the EM algorithm can keep updating until the convergence, and is usually applied to scenarios that miss the observation of implicit variables, such as Gaussian mixture model~\cite{xuan2001algorithms}. 
For 3D reconstruction, we also meet a similar problem to estimate the four reconstruction factors simultaneously. 
Inspired by EM,  we propose a similar optimization strategy in our work, and 
this process actually is a ``Do'' operation in the causal map~\cite{pearl2018book}. 

\section{Method} \label{sec:method}
\subsection{Overview} 
Given a clothing image $I_i$ and the foreground mask $M_i$, we aim to infer the corresponding 3D mesh with texture (see Figure~\ref{fig:pipeline}), where $i \in [1, N]$ and $N$ is the number of the samples in the dataset. We do not require any extra 3D annotations. In this work, 
we explicitly follow the structural causal map in Figure~\ref{fig:causal} (d) to build the whole pipeline. Generally, the spirit of causality helps us to (1) disentangle the 3D attributes from inherent correlations, such as the ambiguity between the shape and camera encoder; (2) re-consider the causal relation between entities, such as the 3D prior template (prototype) and camera position estimation. 
Specifically, we deploy four independent encoders, \ie, \textbf{shape encoder} $E_S$, \textbf{camera pose encoder} $E_C$, \textbf{illumination encoder} $E_A$ and \textbf{texture encoder} $E_{UV}$. 
The decoder is based on the differentiable render~\cite{KaolinLibrary}, which does not contain any learnable parameters. Therefore, we can also regard the render as a fixed decoder. 
Following existing works~\cite{kulkarni2019canonical,li2020umr}, we also introduce a 3D prototype $\bar{S} \in \mathbb{R}^{|\bar{S}|\times3}$, which explicitly involves the prior body structure to the network learning. 
The 3D prior template can be initialized with an arbitrary mesh. Without loss of generality, we apply the elliposid (contains $642$ vertices and $1280$ faces) to initialize the 3D prior template. Here we set $642$ vertices as the default setting to illustrate the proposed approach. 
During training, we keep updating the prior template $\bar{S}$.  When inference, the model 
deploys the latest 3D prior template (prototype). 

\subsection{Model Structure}
\noindent\textbf{Shape Encoder.}
We follow the causal map to explicitly introduce 3D prior into the encoder learning. Given a input image-mask pair $I_i$, $M_i$ and 3D prior template $\bar{S}$, the shape encoder predicts the offsets $\Delta S_i \in \mathbb{R}^{642\times3}$ for every vertex:\vspace{-3mm}
\begin{align}
    \Delta S_i &= E_S(I_i, M_i, \bar{S}), \\
    S_i &= \bar{S} + \Delta S_i. \label{eq:s}\vspace{-2mm}
\end{align}
The final 3D shape $S_i$ is the sum of the 3D prior template and the 3D per-vertex offsets, and $S_i \in \mathbb{R}^{642\times3}$.
Different from most existing works~\cite{hu2021smr,li2020umr}, which independently estimates  $\Delta S_i$ from the input image $I_i$ and the mask $M_i$, our shape encoder explicitly takes the prior template into the deformation prediction as $E_S(I_i, M_i, \bar{S})$. The main idea is straightforward, since predicting shape offsets depends on foreknowing the shape prior. We explicitly provide the shape template to help the training. 
In particular, the shape encoder contains a CNN-based backbone, an integration module (IM) and a fully connected layer (fc). 

\begin{figure}[tbp]
\begin{center}\vspace{-.25in}
    \includegraphics[width=1\linewidth]{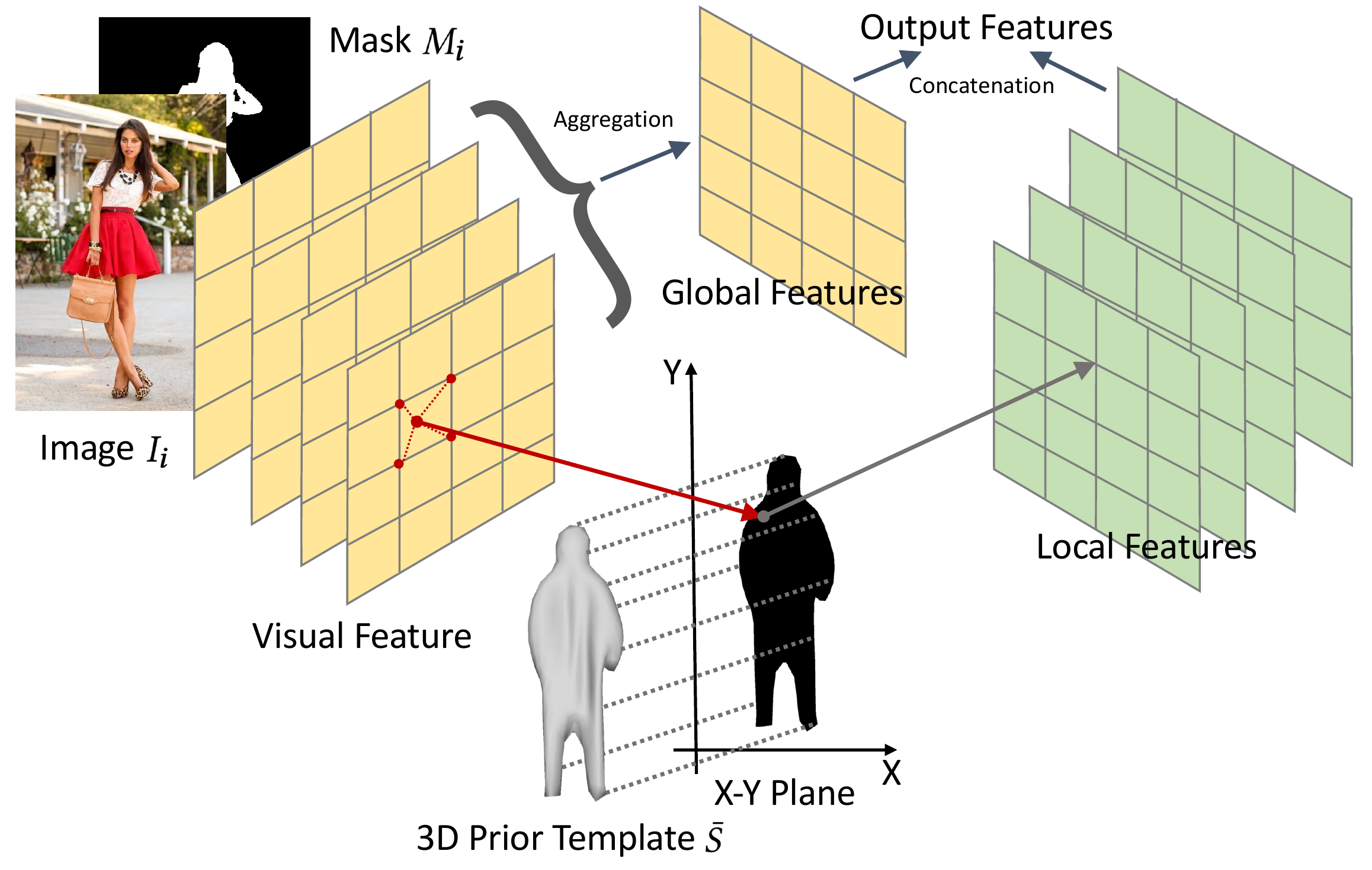}
\end{center}
\vspace{-.3in}
     \caption{ Integration Module (IM). We harness IM to explicitly fuse the prior spatial information from template $\bar{S}$ and the visual feature from $(I_i, M_i)$. The visual feature is the output of the backbone, \eg, HR-Net~\cite{yu2021lite}. As the arrow direction, we leverage the grid sampler to obtain the local feature from the corresponding X-Y of the visual feature. We concatenate global features and local features as final outputs. 
     }\label{fig:shape_enc}
\vspace{-.2in}
\end{figure}

\noindent\textbf{Integration Module.} As shown in Figure~\ref{fig:shape_enc}, we fuse the visual feature from both the input image/mask and the 3D prior template. We harness the integration module (IM) to extract the local visual feature according to the 2D location by projecting the 3D template to the X-Y plane. 
On the other hand, the global feature is generated by averaging the input visual feature (by global average pooling) and then we repeat the aggregated feature as the original size. The final $\Delta S_i$ is predicted by a fully-connected layer on the concatenated feature of both global features and local features. 

\noindent\textbf{Camera Pose Encoder.}
Similarly, the camera pose estimation also depends on the shape prior and the image. However, previous works usually ignore the causal dependency on the shape prior $\bar{S}$. When people infer the object position, \eg,  distance, azimuth, and elevation, it is necessary to foreknow the general object shape (general size, general shape, and symmetry to which axis). Therefore, we also deploy a basic convolutional neural network (CNN) followed by an integration module (IM) and  fully-connected layers as the camera pose encoder, which can be formulated as: \vspace{-2mm} 
\begin{align}
    C_i &= E_C(I_i, M_i, \bar{S}),\vspace{-3mm}
\end{align}
where $C_i$ contains four factors, \ie, distance (1-dim), azimuth (1-dim), elevation (1-dim), and X-Y position offset (2-dim). ``dim'' is dimension. The distance is also formulated as object scale in other works~\cite{li2020umr}, while the azimuth is called as the rotation degree. 
We notice that several existing works~\cite{hu2021smr} do not include X-Y position, since they assume that the object is in the center, but we find that including X-Y position prediction actually improves the camera robustness during both training and testing. (Please see the discussion on camera attribute distribution in experiment.) 

\noindent\textbf{Illumination Encoder.} The illumination encoder is to regress the illumination direction and strength, which can be simply formulated as a 9-channel Spherical Harmonics coefficient~\cite{chen2019dibr}. Therefore, we adopt a basic convolutional neural network followed by a 9-channel fully-connected layer to predict the illumination vector from the input image-mask pair: \vspace{-2mm}
\begin{equation} 
    A_i = E_A(I_i, M_i).\vspace{-2mm}
\end{equation}

\noindent\textbf{Texture Encoder.} In this work, we do not predict the color for every vertex. We follow existing works~\cite{kulkarni2019canonical, kanazawa2018learning} to learn a texture flow as the UV map by a U-Net structure~\cite{ronneberger2015u}. Given the input image $I_i$ and the corresponding foreground mask $M_i$, we first predict the texture flow and then map the color according to the spatial location. \vspace{-2mm}
\begin{equation} 
    UV_i = E_{UV}(I_i, M_i).\vspace{-2mm}
\end{equation}

\noindent\textbf{Decoder (Render).} Finally, the decoder, \ie, render, can reconstruct the 3D mesh with color by simply combining the shape $S_i$ and $UV_i$. If we want to re-project the mesh to the 2D space, we further need the camera pose $C_i$ and the illumination direction $A_i$. Therefore, the reconstructed image $\hat{I_i}$ can be written as:\vspace{-2mm}
\begin{equation} 
    \hat{I_i} = \pi (\phi(S_i, UV_i), C_i, A_i), \vspace{-2mm}
\end{equation}
where $\phi$ is the function to colorize the 3D mesh $S_i$ with the  UV map $UV_i$. $\pi$ denotes the projection function mapping the mesh through camera parameters $C_i$ with the illumination $A_i$. As a side product, we can also obtain the reconstructed foreground mask $\hat{M_i}$ during projection. We note that both $\phi$ and $\pi$ are based on the physical mapping, so there do not contain any learnable parameters.

\subsection{Optimization Objectives}
\begin{figure}[tbp]
\begin{center}\vspace{-.25in}
    \includegraphics[width=1\linewidth]{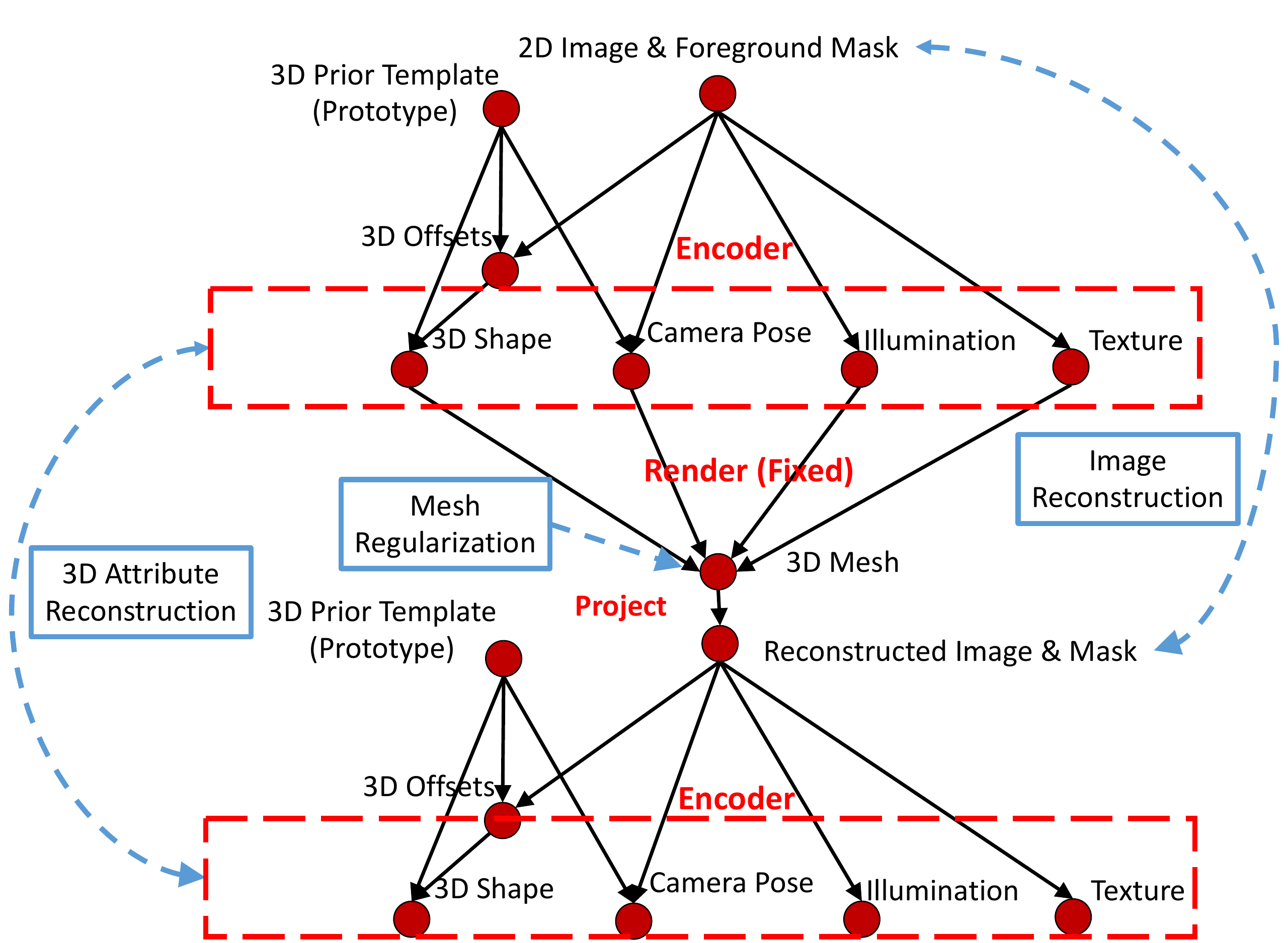}
\end{center}
\vspace{-.25in}
     \caption{ Optimization Objectives. Here we show three kinds of losses on the causal map, which are the image reconstruction loss, the attribute reconstruction loss and the mesh regularization. 
     }\label{fig:loss}
\vspace{-.2in}
\end{figure}

\noindent\textbf{Image Reconstruction Loss.} As shown in the right part of Figure~\ref{fig:loss}, we calculate the pixel level $l_1$ loss of the foreground area between the reconstructed image and the input: \vspace{-4mm}
\begin{equation}
    \mathcal{L}_{img} = \mathbb{E}[|| I_i\odot M_i - \hat{I_i}\odot \hat{M_i} ||_1],
    \vspace{-2mm}
\end{equation}
where $\odot$ denotes element-wise multiplication, and $\mathbb{E}$ denotes the expectation. $\hat{I_i}$ and $\hat{M_i}$ are the reconstructed image and mask projected from the 3D mesh. We note that the $\mathcal{L}_{img}$ focuses on the low-level input. Sometimes the generation quality is good but with small position shifts. To further ensure the generation quality from high-level activations, we also introduce the adversarial loss:  \vspace{-2mm}
\begin{equation}
    \mathcal{L}_{adv} = \mathbb{E}[log D(I_i\oplus M_i) +  log(1-D(\hat{I_i}\oplus \hat{M_i})) ], \vspace{-2mm}
\end{equation}
where $\oplus$ means concatenation. For instance, $I_i\oplus M_i$ is a 4-channel input. $D$ denotes a multi-layer discriminator to classify whether the input is real or generated from our render. In practice, we adopt a basic WGAN structure~\cite{arjovsky2017wasserstein} as the discriminator. Otherwise, we introduce the IoU loss to compare the overlapping area between the generated mask with the ground-truth input mask: \vspace{-3mm}
\begin{equation}
    \mathcal{L}_{IoU} = \mathbb{E}[1-\frac{M_i\cap \hat{M_i}}{M_i\cup \hat{M_i}}].\vspace{-2mm}
\end{equation}

\noindent\textbf{Attribute Reconstruction Loss.}
As shown in Figure~\ref{fig:loss} left part, we also conduct the 3D attribute reconstruction to ensure that the encoder and the decoder are self-consistent:\vspace{-1mm}
\begin{equation}\small{
    \begin{split}
    &\mathcal{L}_{att} = \mathbb{E}[|| S_i - E_S(\hat{I_i}, \hat{M_i},\bar{S}) ||_1] + \mathbb{E}[|| C_i - E_C(\hat{I_i}, \hat{M_i}, \bar{S}) ||_1] \\
    &+ \mathbb{E}[|| A_i - E_A(\hat{I_i}, \hat{M_i}) ||_1] + \mathbb{E}[|| UV_i - E_{UV}(\hat{I_i}, \hat{M_i}) ||_1]. 
    \end{split}} \vspace{-1mm}
\end{equation}
The 3D attributes predicted from the reconstructed image $\hat{I_i}$ should be the same as the predicted attribute from $I_i$. 

\noindent\textbf{Mesh Regularization.}
(1) Laplacian loss~\cite{pixel2mesh} is a regularization to prevent self-intersection of mesh faces. It encourages adjacent vertices to move in the same direction, consequently, avoiding the local part of the mesh producing outrageous deformation. For each vertex position $p$ in the mesh shape $S_i$, the laplacian coordinate is $\delta_p = p - \sum_{k\in K(p)}\frac{k}{||K(p)||}$, where $K(p)$ is the neighbor vertices of $p$ with connected edges.
Specifically, the laplacian loss can be defined as $\mathcal{L}_{lpl} = \mathbb{E}[||\hat{\delta_p}-\delta_p||^2_2$], where $\hat{\delta_p}$ and $\delta_p$ are laplacian coordinates of a vertex before and after the updation respectively; 
(2) Flatten loss is another regularization for keeping faces from intersecting~\cite{pixel2mesh}. The cosine of the angle between two adjacent faces is calculated. The flatten loss is defined as $\mathcal{L}_{flat} = \mathbb{E}[(cos(\Delta\theta_i)+1)^2]$, where  $\Delta\theta_i$ is the angle between two adjacent faces. The angle around $180^{\circ}$ implies a smooth mesh surface; 
(3) Symmetry loss constrains mesh deformations to be reflectional symmetric in the depth~\cite{tulsiani2020implicit}. 
It can be expressed as $\mathcal{L}_{sym} = \mathbb{E}[||Z(p) + Z(\tilde{p})||_1$], where $Z$ denotes the depth of the vertex and $\tilde{p}$ is the reflected vertex of $p$;
(4) Deformation loss~\cite{kanazawa2018learning,li2020umr} is a regularization to prevent the mesh from deforming excessively and  facilitate the average shape learning: $\mathcal{L}_{deform} = \mathbb{E}[||\Delta S||_2$].

\noindent\textbf{Total Loss.} We train four encoders and the discriminator to optimize the total objective, which is a weighted sum of above-mentioned losses:\vspace{-2mm}
\begin{equation} \small{
    \begin{split} 
    &\mathcal{L}_{total} = \lambda_{rec}(\mathcal{L}_{img} + \mathcal{L}_{IoU})  + \lambda_{att}\mathcal{L}_{att}+ \lambda_{adv}\mathcal{L}_{adv} \\ 
    &+  \lambda_{reg}( \mathcal{L}_{sym} + \mathcal{L}_{deform} +  \lambda_{lpl}\mathcal{L}_{lpl} + \lambda_{flat}\mathcal{L}_{flat}).
    \end{split}}\vspace{-2mm}
\end{equation}
In practice, we refer to existing works~\cite{pixel2mesh, hu2021smr, li2020umr} and empirically set $ \lambda_{rec} = 2$, $\lambda_{att} = 1$, $\lambda_{adv} = 1\times10^{-5}$, $\lambda_{reg} = 0.1$, $\lambda_{lpl} = 0.1$, and $\lambda_{flat} = 0.01$. 

\subsection{Optimization Strategy}~\label{sec:strategy}
\noindent\textbf{Following the causality-aware design, two loops are introduced during optimization as the causal ``intervention'' tools.} While estimating one cause, the relations between the outcome and the other ``colliders'' are cut off. In this way, the \textbf{encoder loop} helps to disentangle the correlated relations between the four 3D attributes $S_i, UV_i, C_i, A_i$, while the \textbf{prototype loop} separates the prototype updating $\bar{S}$ from the shape offset estimation $\Delta S_i$.

\noindent\textbf{Encoder Loop.} During the implementation, we notice one main challenge is simultaneously optimizing the four encoders. The problem mainly lies in the image reconstruction loss. For instance, even if three out of four encoders provide correct prediction, the rest provides the wrong attribute, such as incorrect shape offsets. All four encoders are penalized equally. This is one typical ``collider'' case in causality. Therefore, one straightforward idea is to train one encoder (predict one cause) while fixing the other three encoders (control other causes). In this way, we can effectively penalize the target encoder. In particular, we adopt one expectation-maximization loop, which also is an ``invention'' tool in causal learning. For example, we fix the three encoders, \eg, $E_c$, $E_A$, $E_{UV}$, to cut the arrows from inputs to three attributes, \ie, $C_i$, $A_i$ and $UV_i$.  Only the shape attribute $S_i$ still keeps the dependency from the input image-mask pair. Hence, when the loss is back-propagating, only the shape encoder is penalized. In this way, we disentangle four encoders not only in the forward passing design (independent encoder weights) but also in the loss of back-propagation.

\noindent\textbf{Prototype Loop.} 
We also observe a common ambiguity between prototype updating and shape estimation. The problem is mostly due to Eq.~\ref{eq:s}. Since it is an addition equation, during gradient back-propagation, $\bar{S}$ and $\Delta S_i$ receive the punishment equally. It is hard to distinguish $\bar{S}$ from $\Delta S_i$. 
Therefore, we adopt the causal invention tool, \ie, Expectation-Maximization, again. During training, we fix the prototype (control one cause) and maximize the shape offsets likelihood (predict another cause). After every training epoch, we leverage the mean shape offsets to update $\bar{S} = \bar{S} + \mathbb{E}[\Delta S]$. In practice, different from existing works (\eg, two-stage training~\cite{li2020umr}), we adopt a linear warming-up strategy~\cite{goyal2017accurate} to update prototype slowly in the early epochs and harness the exception handling by clipping extreme deformations. In this way, we disentangle the prototype updating from the shape offsets estimation and learn the model in one go.

\section{Experiment} \label{sec:exp}
We evaluate the proposed approach on two fashion datasets, \ie, ATR~\cite{ATR} and Market-HQ ~\cite{zheng2015scalable}, and a widely-used bird dataset CUB~\cite{WahCUB_200_2011}. 
Since there are no ground-truth 3D meshes, we follow existing works~\cite{hu2021smr} and adopt 2D metric, \ie, FID~\cite{heusel2017gans}, SSIM~\cite{wang2004image}, and MaskIoU, to evaluate the ``2D $\rightarrow$ 3D $\rightarrow$ 2D'' process. FID compares the distribution of two sets of images. We denote the 3D reconstruction results as FID$_{recon}$, the generated image with different viewpoints as FID$_{novel}$, following \cite{hu2021smr}. For Market-HQ, we also report FID$_{90}$ by comparing generated side-view images with real images. Please see \textbf{supplementary material} for dataset preparation, structure and training details. 

\setlength{\tabcolsep}{10pt}
\begin{table}\vspace{-.2in}
\caption{Comparison with two off-the-shelf template-based  methods on the human clothing ATR dataset. Since no texture mapping is contained in the template-based methods, we only compare MaskIoU (\%), which reflects the ``2D $\rightarrow$ 3D $\rightarrow$ 2D'' reconstruction quality on the unseen test set. 
}\vspace{-.25in}
\label{table:ATR}
\begin{center}\footnotesize
\begin{tabular}{c|c|c|c}
\hline
Methods &    HMR~\cite{kanazawaHMR18}    & ROMP~\cite{ROMP} & Ours  \\
\shline
MaskIoU (\%) $\uparrow$ & 69.7  & 70.3 & 81.1 \\
\hline
\end{tabular} \vspace{-.25in}
\end{center}
\end{table}

\begin{figure*}[tbp]
\centering \vspace{-.25in}
\animategraphics[autoplay,loop, width=1\linewidth]{10}{images/temp/rainbow_}{1}{36}
\vspace{-.3in}
  \caption{Novel-view 3D clothing generation from single images on the unseen test set of Market-HQ and ATR. 
  {\color{citecolor}(Please open the paper in Adobe Reader to see the mesh rotation.)} Here we gradually ``Do'' / change the camera azimuth degree to render the human.
}
\vspace{-.2in}
\label{fig:multi_view}
\end{figure*}

\subsection{Quantitative Experiments}
\noindent\textbf{Comparison with Template-based Methods.} 
We first compare with the off-the-shelf template-based methods~\cite{kanazawaHMR18,ROMP} in Table~\ref{table:ATR}. This line of methods is based on the body template with great structure robustness, but is not well scalable to the non-rigid clothing. Since no texture mapping function is built in the template-based methods, we focus on comparing MaskIoU (\%), which reflects the ``2D $\rightarrow$ 3D $\rightarrow$ 2D'' shape reconstruction quality. We observe that the proposed method achieves a higher MaskIoU score of $81.1\%$ on the test set. The result is also consistent with the visualization in Figure~\ref{fig:motivation}. For clothing reconstruction, the proposed method is more scalable than the template-based methods, covering more regions of interest.

\noindent\textbf{Comparison with Single-image Reconstruction Methods.} As shown in Table~\ref{table:methods_comparison}, we compare the proposed method with other state-of-the-art approaches~\cite{hu2021smr,li2020umr,kanazawa2018learning,chen2019dibr,li2020online,bhattad2021view,ye2021shelf} on CUB. MeshInversion~\cite{zhang2022monocular} deploys test time optimization, so here we do not include it. Among the existing works, SMR~\cite{hu2021smr} has achieved the high-fidelity reconstruction and novel-view generation performance. In contrast, the proposed method yields a better reconstruction performance ($81.8\%$ MaskIoU and $83.5\%$ SSIM). 
At the meantime, for novel-view generation, our method also has achieved 63.5 FID$_{novel}$, surpassing SMR by a clear margin.  Similarly, based on the same backbone, our method also surpasses SMR on Market-HQ and ATR (see Table ~\ref{table:ablation}). 

\setlength{\tabcolsep}{5pt}
\begin{table}
\caption{Comparison with other single-image reconstruction methods on the CUB bird dataset. MaskIoU (\%) and  SSIM (\%) reflects the front-view reconstruction quality on the unseen test set, while FID$_{novel}$ compares the distribution difference between generated images from novel views and the original dataset.
}\vspace{-.2in}
\label{table:methods_comparison}
\begin{center} \footnotesize
\begin{tabular}{c|c|c|c}
\hline 
Methods & MaskIoU (\%) $\uparrow$       & SSIM   (\%) $\uparrow$     &  FID$_{novel}$ $\downarrow$  \\
\shline
View-gen~\cite{bhattad2021view} & 61.7 & - & 70.3 \\
 ShSMesh~\cite{ye2021shelf} & 70.7 & - & 161.0 \\
 CMR~\cite{kanazawa2018learning} & 73.8 & 44.6 & 115.1 \\
 UMR~\cite{li2020umr}            & 73.4 & 71.3 & 83.6  \\
 DIB-R~\cite{chen2019dibr}       & 75.7 & -     & -     \\
 ACMR-vid~\cite{li2020online} & 77.3 & -     & -     \\
 SMR~\cite{hu2021smr} & 80.6 & 83.2 & 79.2 \\
 \hline
 Ours & \textbf{81.8}  & \textbf{83.5} & \textbf{63.5} \\
\hline
\end{tabular}  \vspace{-.2in}
\end{center}
\end{table}

\setlength{\tabcolsep}{2pt}
\begin{table}
\caption{Ablation Study on Market-HQ and ATR. 
``No IM'' denotes that we remove the integration module. We observe that although ``No Encoder Loop'' leads the model to over-fit the front-view reconstruction FID$_{recon}$, the side-view  performance FID$_{90}$ is extremely poor. In contrast, our full model takes a balance point between reconstruction, \ie, MaskIoU, and novel-view generation, \ie, FID$_{novel}$. (Considering the majority of ATR test set is close-frontal view, we do not report FID$_{90}$ on ATR.)
}\vspace{-.1in}
\label{table:ablation}\vspace{-.15in}\footnotesize
\begin{center}\resizebox{\linewidth}{!}{
\begin{tabular}{c|c|c|c|c|c|c|c|c|c}
\hline 
\multirow{3}{*}{Methods} & \multicolumn{5}{c|}{Market-HQ} & \multicolumn{4}{c}{ATR}\\
 & MaskIoU        & SSIM            & \multirow{2}{*}{FID$_{recon}$  $\downarrow$}    & \multirow{2}{*}{FID$_{novel}$ $\downarrow$} &  \multirow{2}{*}{FID$_{90}$ $\downarrow$} & MaskIoU        & SSIM            & \multirow{2}{*}{FID$_{recon}$  $\downarrow$}    & \multirow{2}{*}{FID$_{novel}$ $\downarrow$} \\
& (\%) $\uparrow$ & (\%) $\uparrow$ &  &  &  & (\%) $\uparrow$ & (\%) $\uparrow$ &  &  \\ 
\shline
SMR$^\dagger$ & 81.0 & 66.1 & 23.6 & 60.0 & 120.5 & 78.5 & 72.9 & 38.5 & 76.7 \\ 
No IM & 72.0 & 56.4 & 44.6 & 72.5 & 107.7 & 77.3 & 72.4 & 43.0 & 81.6 \\ 
No Prototype Loop & 82.6 & 65.8 & 21.9 & 47.2 & 104.1 & 80.7 & 71.9 & 37.3 & 72.2 \\ 
No Encoder Loop & 82.9 & \textbf{67.9} & \textbf{17.4} & 49.2 &  {\color{red}176.0} & 76.7 & 71.6 & \textbf{33.1} & 67.0 \\ 
 \hline
Ours & \textbf{83.4} & 66.3 & 21.5 & \textbf{46.7} & \textbf{93.3} & \textbf{81.1} & \textbf{72.6} & 35.9 & \textbf{66.8} \\ 
\hline
\end{tabular}}
\end{center}\vspace{-.1in}
\footnotesize{$^\dagger$: For a fair comparison, we re-implement SMR with the same backbone as ours and enable XY position prediction. 
}
\vspace{-.2in}
\end{table}

\subsection{Qualitative Experiments} 
\noindent\textbf{Reconstruction and Novel-view Results.}
As shown in Figure~\ref{fig:multi_view}, we reconstruct the person with non-rigid clothing. 
We could observe that the model not only successfully learns legs and arms, but also captures non-rigid objects, including hair, dress, and handbag. CUB results are in Fig.~\ref{fig:cub}.  

\noindent\textbf{Exchanging Clothing.}
Inspired by 2D GAN-based work~\cite{zheng2019joint}, we also show the result of changing the texture of any two persons but with a 3D mesh manner (see Figure~\ref{fig:rainbow}). In particular, we apply the shape encoder and the texture encoder to extract the shape $S_i$ and the UV texture map $UV_j$, respectively. Then we deploy the render to generate the new mesh based on $S_i$ and $UV_j$. The first row and the first column are the input RGB images. The rest is the projected results of the new 3D meshes. We rotate the mesh for better 3D visualization. It verifies the robustness of our method. The learned UV map could be successfully aligned to different human meshes, even though we have not introduced any part annotations during the training process. 

\begin{figure*}[tbp]
\centering \vspace{-.25in}
\animategraphics[autoplay,loop, width=1\linewidth]{10}{images/camera/rainbow_}{1}{12}
\vspace{-.25in}
  \caption{Novel-view 3D clothing generation from single images on the unseen test set of Market-HQ. Here we gradually ``Do'' / change the camera distance, elevation and XY-position to render the human. {\color{citecolor}(Please open the paper in Adobe Reader to see the movement.)}
} \vspace{-.2in}
\label{fig:camera}
\end{figure*}

\begin{figure}[tbp]
\centering
\animategraphics[autoplay,loop, width=0.7\linewidth]{10}{images/rainbow/rainbow_}{1}{36}
\vspace{-.1in}
  \caption{3D clothing changing by exchanging the 3D mesh shape and texture. {\color{citecolor}(Please open Adobe Reader to see the movement.)}
}
\label{fig:rainbow}
\vspace{-.2in}
\end{figure}

\noindent\textbf{Manipulate Camera Attributes.}
Since four encoders are disentangled, the proposed method could easily manipulate 3D attributes for customization. Besides the rotation (interpolating the azimuth degree), we could also leverage the learned model to change distance, elevation, and XY position. As shown in Figure~\ref{fig:camera}, we could observe that the proposed method successfully disentangles these camera parameters and could control the projected result smoothly.

\subsection{Ablation Study and Further Discussion} 

\noindent\textbf{Does the two expectation-maximization loops help the encoder learning?} Yes. As shown in Table~\ref{table:ablation}, we conduct two ablation studies on Market-HQ and ATR. (1) One is to stop the prototype updating, \ie, No Prototype Loop, and we deploy the fixed elliposid as the basic shape. It directly limits the shape deformation, compromising the reconstruction performance.
(2) Besides, we also explore training all the encoders simultaneously without the encoder loop as ``No Encoder Loop''. We observe that the model can easily over-fit the front-view reconstruction quality but it does not perform well in the novel view, especially when we look at the 3D mesh from the side view, \ie, $90^{\circ}$.  

\noindent\textbf{Does the integration module work?} Yes. We design the integration module to explicitly fuse the prior prototype as local features for learning shape. Removing the integration module, \ie, No IM, leads to a performance drop in both reconstruction and novel-view generation (see Table~\ref{table:ablation}). 

\noindent\textbf{Person Re-id.} 
One interesting problem remains whether our learned 3D human model can facilitate downstream tasks, such as person re-id, which intends to match the pedestrian from different viewpoints. We do not intend to pursue state-of-the-art performance, but verify the relative improvement of using the generated data (see Table~\ref{table:baseline}). 
We observe that our generated 3D-aware images can facilitate re-id representation learning. More details are in \textbf{suppl}.

\noindent\textbf{Camera Attribute Distribution.} We observe that the camera encoder successfully captures the camera distribution in the Market-HQ test set, which is aligned with the dataset collection setting. Please check \textbf{suppl.} for details. 

\noindent\textbf{Limitations.} There are two faces or two backs of heads on one reconstructed mesh, commonly called Janus Issue. 
It is because our work is still based on a single image, and the learned model only ``sees'' one single view of the human. Especially on ATR (most photos are  close-frontal faces), it can not learn the 3D prior, \ie, one person only has one face. Therefore, even if we introduce WGAN discriminator~\cite{arjovsky2017wasserstein}, it can not provide 3D-aware adversarial loss. The model still largely relies on the symmetric structure to generate the back view. Hence, we think that, in the future, the large-scale multi-view image datasets~\cite{EVA3D} may help to further solve this limitation upon our work. We  
also tried simply replacing the ellipsoid with an SMPL template~\cite{SMPL:2015}, but it fails due to the optimization problem on too many vertices \& initial arm position (see \textbf{suppl.} discussion).  

\setlength{\tabcolsep}{5pt}
\begin{table}
\caption{ The re-id performance improvement on Market-1501. 
We train two competitive backbones. The results suggest that our generated 3D-aware data can further facilitate representation learning. 
}
\label{table:baseline}\vspace{-.2in}
\begin{center}\footnotesize
\begin{tabular}{c|c|c|c}
\hline 
Methods &  Training Set     & Rank@1  &  mAP \\
\shline
\multirow{2}{*}{ResNet50-ibn~\cite{pan2018IBNNet}} & Original  & 94.63 & 87.37 \\
 & Original+3D  & 95.07	& 87.80 \\
\hline
\multirow{2}{*}{HR18-Net~\cite{wang2020deep}} & Original  & 94.74 & 88.13 \\
 & Original+3D  & 95.43 & 88.54 \\
\hline
\end{tabular}
\end{center}\vspace{-.2in}
\end{table}

\begin{figure}[tbp]
\begin{center}
    \animategraphics[autoplay,loop, width=0.95\linewidth]{10}{images/cub/cub_}{1}{36}
\end{center}
\vspace{-.3in}
     \caption{ Novel-view 3D bird generation on the test set of CUB. 
     }\label{fig:cub}
\vspace{-.2in}
\end{figure}

\section{Conclusion}\label{sec:conclusion}
In this paper, we study the 3D clothing reconstruction task to build a ``3D Magic Mirror''. 
We follow the spirit of the structural causal map to re-design the output dependency, and leverage two expectation-maximization loops to facilitate the training process. 
Despite using relatively weak supervision, the proposed method is still competitive with other existing works, and shows great scalability to different non-rigid objects.  In the future, we will further explore the applications to multi-modality generation~\cite{xue2022ulip} and 3D object re-id~\cite{sun2019dissecting,han20223d,zheng2020parameter}. 

{\small
\bibliographystyle{ieee_fullname}
\bibliography{egbib}
}

\clearpage

\appendix

\setcounter{table}{5}
\setcounter{figure}{10}

\section{Dataset Preparation} 
\noindent\textbf{ATR.} ATR is a large-scale fashion dataset~\cite{ATR}. It contains $17,700$ human body images, with 18 detailed semantic annotations for every image. The dataset is split into $16,000$ training images, $700$ validation images, and $1,000$ test images. In this work, we do not use any part segmentation annotations but only leverage the binary foreground mask as the weak annotation. 
It is because foreground masks are easy to obtain, which is closer to our application in real-world scenarios. 

\noindent\textbf{Market-HQ.} We build a high-resolution variable of the Market-1501 dataset~\cite{zheng2015scalable} based on academic usage. The original Market-1501 is with a relatively low resolution of $128\times64$. We first apply Real-ESRGAN~\cite{wang2021realesrgan} to upsample the images to $512\times256$. 
To acquire the foreground mask, we have tried the off-the-shelf human parsing model~\cite{ruan2019devil, luo2018macro}, but such models suffer from the dataset domain gap. Instead, we apply HMR~\cite{kanazawaHMR18} to obtain the pseudo foreground mask via 2D projection. 
As a result, Market-HQ contains $12,936$ images of $751$ persons for training and $3,386$ test images of other $750$ persons. 
There is no overlapping human appearing in both the training and test set. 

\noindent\textbf{CUB-200-2011.} CUB-200-2011 is one of the prevailing datasets for single-view 3D reconstruction tasks~\cite{WahCUB_200_2011}. It contains images from 200 subcategories of birds, of which $5,994$ are for training and $5,794$ are for testing. 2D foreground masks and keypoints are provided. In this work, we only deploy 2D foreground masks.

\section{Implementation Details} 
Our model consists of four encoders, one render, and one discriminator. Four encoders and the discriminator are implemented based on Pytorch~\cite{NEURIPS2019}, and the differentiable render is from Kaolin~\cite{KaolinLibrary}. In the following part, we use channel $\times$ height $\times$ width to indicate the size of feature maps. 
For training Market-HQ, all input images are resized to $128\times64$ from a high resolution and the image quality is still better than that of images in the original dataset. 
For ATR, all input image is resized as $160\times96$.
For  CUB Bird, all images are padded to a square input and then resized to $128\times128$. 
During the evaluation, some generated results are up-sampled to $256\times256$ for a fair comparison. 
Random horizontal flipping is used for data augmentation. We apply Adam~\cite{kingma2014adam} to optimize the encoders with a mini-batch of 48 and set the basic learning rate as $5\times10^{-5}$, and $(\beta_1, \beta_2) = (0.95, 0.99)$. The number of training epochs is set as $600$. 
Besides, since the backbone model, \ie, HRNet-Lite~\cite{yu2021lite}, has been pre-trained on ImageNet, we do not intend the backbone to update too fast and set the learning rate of the backbone as $0.05 \times $ the basic learning rate. 
We also modified the first layer of HRNet-Lite for 4-channel input (RGB image + foreground mask). In particular, we apply the mean filter weights of the original first convolutional layer to initialize the fourth channel filters. Next, we illustrate the network architecture in detail. 

\section{Network Architectures}
\label{sec:architecture}
The proposed method consists of the camera encoder $E_C$, shape encoder $E_S$, texture encoder $E_{UV}$, illumination encoder $E_A$ and discriminator $D$. 
Following the common practice in GANs, we mainly adopt convolutional layers and residual blocks \cite{he2016deep} to construct them. The model can be applied to different input scales. Taking the Market dataset as an example, we utilize the input size as $4\times128\times64$ for illustration.  
MMPool denotes the gem pooling~\cite{radenovic2018fine}, which is a weighted sum of average pooling and max pooling. 

(1) Shape Encoder: Table \ref{tab:es} shows the architecture of $E_s$. We apply the backbone network, \ie, HRNet-Lite-v2~\cite{yu2021lite}, to extract the visual feature. Then we apply the Integration Module (IM) to fuse the visual feature and 3D prior template. In the IM block, we concatenate the local feature (2048-dim), global feature (2048-dim), neighbor difference feature (2048-dim), and the template coordinate (3-dim), so the output size of the IM is 6147 dimensions. In practice, we simply subtract the local feature of every vertex with the mean neighbor feature as the neighbor difference feature, where the mean neighbor feature is the mean local feature of the connected vertex.
After each convolutional layer, we generally apply the batch normalization layer and LReLU (negative slope set to 0.2). 
Finally, we obtain 1926-dim ($1926 = 642\times3$) output, which is the XYZ biases for 642 vertices as $\Delta S$.

(2) Camera Position Encoder: 
Table \ref{tab:ec} shows the architecture of encoder $E_c$. 
We deploy residual blocks and convolutional layers to build the model. 
For better location estimation, we follow CoordConv~\cite{liu2018intriguing} and concatenate the grid. 
ResBlock$\_$half denotes the downsampling block with residual connection from ~\cite{he2016deep}. Since we concatenate the downsampled input, the output size of ResBlock$\_$half is doubled. 
The final IM simply concatenates the local feature and global feature, so the output size is 576 $\times$ 2 $\times$ 2. We then flatten the feature and apply three independent MLPs (each contains 2 fully-connected layers) for $azimuths_y$ \& $azimuths_x$ (2-dim), elevation \& distance (2-dim), and XY-biases (2-dim). The final azimuth is arctan($azimuths_y / azimuths_x$), which is one dimension.  

(3) Illumination Encoder: As shown in Table~\ref{tab:ea}, we deploy one convolutional neural network to predict the illumination (9-dim). We apply both the batch normalization layer and LReLU (negative slope set to 0.2) after every convolutional layer.

(4) Texture Encoder: We deploy a U-Net structure in \cite{hu2021smr} and adopt light-weight ResNet34~\cite{he2016deep} as U-Net encoder. 
We also adopt the BiFPN structure~\cite{tan2020efficientdet} to facilitate the communication between different layers when decoding.
Due to the limitation of the table, we could not show the skip connections and the entire up-sampling process of decoder.  
Therefore, here we only show the main components in Table~\ref{tab:euv}. The final UV map is symmetric, so we concatenate the mirrored output.   

(5) Discriminator: We deploy one convolutional neural network to obtain the real/fake prediction (see Table~\ref{tab:d}). We only apply LReLU (negative slope set to 0.2) after every convolutional layer. 


\begin{table}[tbp]
\caption{Architecture of the shape encoder $E_S$.} 
\label{tab:es}\resizebox{\linewidth}{!}{\footnotesize
{
\setlength{\tabcolsep}{10pt}
\begin{tabular}{l|c|c}
\shline
Layer & Parameters & Output Size \\
\hline 
Input                    & -          & 4 $\times$ 128 $\times$ 64 \\
Backbone & 16 M & 2048 $\times$ 4 $\times$ 2 \\
(HRNet-Lite-v2) & & \\
\hline
IM Module                &  -       & 6147 $\times$ 642 $\times$ 1 \\
Conv1d                   & \convblock{256}{1}  & 256 $\times$ 642 $\times$ 1 \\
Conv1d                   & \convblock{3}{1}    & 3   $\times$ 642 $\times$ 1\\
\hline
FC                       & [1926, 1926]        & 1926 \\
\shline
\end{tabular}}}
\end{table}

\begin{table}[tbp]
\caption{Architecture of the camera position encoder $E_C$.} 
\label{tab:ec}\resizebox{\linewidth}{!}{\footnotesize
{
\setlength{\tabcolsep}{10pt}
\begin{tabular}{l|c|c}
\shline
Layer & Parameters & Output Size \\
\hline 
Input         & -                 & 4 $\times$ 128 $\times$ 64 \\
\hline
AddCoords2d   & -            & 6 $\times$ 128 $\times$ 64 \\
Conv          & \convblock{36}{5} & 36 $\times$ 64 $\times$ 32 \\
\multirow{3}{*}{ResBlock\_half} & \resblock{36}{1} & \multirow{3}{*}{ 72 $\times$ 32 $\times$ 16} \\
  &  &  \\
  &  &  \\
\multirow{3}{*}{ResBlock} & \resblock{72}{1} & \multirow{3}{*}{ 72 $\times$ 32 $\times$ 16} \\
  &  &  \\
  &  &  \\
\hline
\multirow{3}{*}{ResBlock\_half} & \resblock{72}{1} & \multirow{3}{*}{ 144 $\times$ 16 $\times$ 8} \\
  &  &  \\
  &  &  \\
\multirow{3}{*}{ResBlocks} & \resblock{144}{3} & \multirow{3}{*}{ 144 $\times$ 16 $\times$ 8} \\
  &  &  \\
  &  &  \\
\hline
\multirow{3}{*}{ResBlock\_half} & \resblock{144}{1} & \multirow{3}{*}{ 288 $\times$ 8 $\times$ 4} \\
  &  &  \\
  &  &  \\
\multirow{3}{*}{ResBlocks} & \resblock{288}{6} & \multirow{3}{*}{ 288 $\times$ 8 $\times$ 4} \\
  &  &  \\
  &  &  \\
\hline
IM & - & 576 $\times$ 8 $\times$ 4  \\
MMPool & - & 576 $\times$ 2 $\times$ 2   \\
\multirow{2}{*}{MLPs $\times$ 3} & \multirow{2}{*}{\(\left[\begin{array}{c}\text{2304, 128}\\[-.1em] \text{128, 2} \end{array}\right]\)$\times$3} & 128 $\times$ 3 \\
 & & 2 $\times$ 3 \\
\shline
\end{tabular}}}
\end{table}

\begin{table}[t]
\caption{Architecture of the illumination encoder $E_A$.} 
\label{tab:ea}\resizebox{\linewidth}{!}{\footnotesize
{
\setlength{\tabcolsep}{10pt}
\begin{tabular}{l|c|c}
\shline
Layer & Parameters & Output Size \\
\hline 
Input         & -                 & 4 $\times$ 128 $\times$ 64 \\
\hline
Conv          & \convblock{32}{5} & 32 $\times$ 64 $\times$ 32 \\
Conv          & \convblock{64}{5} & 64 $\times$ 32 $\times$ 16 \\
Conv          & \convblock{96}{5} & 96 $\times$ 16 $\times$ 8 \\
Conv          & \convblock{192}{5} & 192 $\times$ 8 $\times$ 4 \\
Conv          & \convblock{96}{5} & 96 $\times$ 4 $\times$ 2 \\
MMPool        & - & 96 $\times$ 1 $\times$ 1   \\
\hline
FC                       & [96, 48]        & 48 \\
\hline
FC                       & [48, 9]        & 9 \\
\shline
\end{tabular}}}
\end{table}

\begin{table}[tbp]
\caption{Architecture of the texture encoder $E_{UV}$.} 
\label{tab:euv}\resizebox{\linewidth}{!}{\footnotesize
{
\setlength{\tabcolsep}{10pt}
\begin{tabular}{l|c|c}
\shline
Layer & Parameters & Output Size \\
\hline 
Input         & -                 & 4 $\times$ 128 $\times$ 64 \\
\hline 
Conv (ResNet-34)         & \convblock{64}{7} & 64 $\times$ 64 $\times$ 32 \\
MaxPooling    &  & 64 $\times$ 32 $\times$ 16 \\
\multirow{3}{*}{ResBlock (ResNet-34)} & \resblock{64}{3} & \multirow{3}{*}{ 64 $\times$ 32 $\times$ 16} \\
  &  &  \\
  &  &  \\
\hline
\multirow{3}{*}{ResBlocks (ResNet-34)} & \resblock{128}{4} & \multirow{3}{*}{ 128 $\times$ 16 $\times$ 8} \\
  &  &  \\
  &  &  \\
\hline
\multirow{3}{*}{ResBlocks (ResNet-34)} & \resblock{256}{6} & \multirow{3}{*}{ 256 $\times$ 8 $\times$ 4} \\
  &  &  \\
  &  &  \\
\hline
\multirow{3}{*}{ResBlock (ResNet-34)} & \resblock{512}{3} & \multirow{3}{*}{ 512 $\times$ 4 $\times$ 2} \\
  &  &  \\
  &  &  \\
\hline
Conv          & \convblock{256}{3} & 256 $\times$ 4 $\times$ 2 \\
\multirow{3}{*}{ResBlock} & \resblock{256}{1} & \multirow{3}{*}{ 256 $\times$ 4 $\times$ 2} \\
  &  &  \\
  &  &  \\
Upsample     & - & 256 $\times$ 8 $\times$ 4 \\
\hline
Conv          & \convblock{128}{3} & 128 $\times$ 8 $\times$ 4 \\
\multirow{3}{*}{ResBlock} & \resblock{128}{1} & \multirow{3}{*}{ 128 $\times$ 8 $\times$ 4} \\
  &  &  \\
  &  &  \\
Upsample     & - & 128 $\times$ 16 $\times$ 8 \\
\hline
Conv          & \convblock{64}{3} & 64 $\times$ 16 $\times$ 8 \\
\multirow{3}{*}{ResBlock} & \resblock{64}{1} & \multirow{3}{*}{ 64 $\times$ 16 $\times$ 8} \\
  &  &  \\
  &  &  \\
Upsample     & - & 64 $\times$ 32 $\times$ 16 \\
\hline
Conv          & \convblock{64}{3} & 64 $\times$ 32 $\times$ 16 \\
\multirow{3}{*}{ResBlock} & \resblock{64}{1} & \multirow{3}{*}{ 64 $\times$ 32 $\times$ 16} \\
  &  &  \\
  &  &  \\
Upsample     & - & 64 $\times$ 64 $\times$ 32 \\
\hline
Conv          & \convblock{32}{3} & 32 $\times$ 64 $\times$ 32 \\
\multirow{3}{*}{ResBlock} & \resblock{32}{1} & \multirow{3}{*}{ 32 $\times$ 64 $\times$ 32} \\
  &  &  \\
  &  &  \\
Upsample      & - & 32 $\times$ 128 $\times$ 64 \\
\hline
Conv          & \convblock{2}{3} & 2 $\times$ 128 $\times$ 64 \\
Tanh & - & 2 $\times$ 128 $\times$ 64 \\
Project & Grid Sampler & 3 $\times$ 128 $\times$ 64 \\
Concat  & Flipping & 3 $\times$ 256 $\times$ 64 \\
\shline
\end{tabular}}}
\end{table}

\begin{table}[t]
\caption{Architecture of the discriminator $D$.}
\label{tab:d}\resizebox{\linewidth}{!}{\footnotesize
{
\setlength{\tabcolsep}{8pt}
\begin{tabular}{l|c|c}
\shline
Layer & Parameters & Output Size \\
\hline 
Input & - & 4 $\times$ 128 $\times$ 64 \\
\hline
Conv1 & \convblock{16}{1} & 16 $\times$ 128 $\times$ 64 \\
Conv2 & \convblock{16}{3} & 16 $\times$ 128 $\times$ 64 \\
Conv3 & \convblock{32}{3}  & 32 $\times$ 64 $\times$ 32 \\
\hline
Conv4 & \convblock{32}{3}  & 32 $\times$ 64 $\times$ 32 \\
Conv5 & \convblock{48}{3}  & 48 $\times$ 32 $\times$ 16 \\
\hline
Conv6 & \convblock{48}{3}  & 48 $\times$ 32 $\times$ 16 \\
Conv7 & \convblock{64}{3}  & 64 $\times$ 16 $\times$ 8 \\
\hline
Conv8 & \convblock{64}{3}  & 64 $\times$ 16 $\times$ 8 \\
Conv9 & \convblock{64}{3}  & 64 $\times$ 8 $\times$ 4 \\
\hline
Conv10 & \convblock{64}{3}  & 64 $\times$ 8 $\times$ 4 \\
Conv11 & \convblock{64}{3}  & 64 $\times$ 4 $\times$ 2 \\
\hline
Conv12 & \convblock{64}{3}  & 64 $\times$ 4 $\times$ 2 \\
Conv13 & \convblock{48}{3}  & 48 $\times$ 2 $\times$ 1 \\
\hline
Conv14 & \convblock{32}{3}  & 32 $\times$ 2 $\times$ 1 \\
Conv15 & \convblock{1}{1}   & 1 $\times$ 2 $\times$ 1 \\
Mean   & - & 1 \\
\shline
\end{tabular}}}
\end{table}

\setlength{\tabcolsep}{3pt}
 \begin{table}[tbp]
\caption{Architecture of the IM Module.} 
\label{tab:im}\resizebox{\linewidth}{!}{\footnotesize
{
\setlength{\tabcolsep}{10pt}
\begin{tabular}{l|l|c|c}
\shline
Data & Layer   & Parameters & Output Size \\
\hline 
Input Feature &                   & -          & 2048 $\times$ 4 $\times$ 2 \\
Template &      & - & \textbf{3} $\times$ 642 $\times$ 1   \\
\hline 
Local & Grid Sampler  & -          & \textbf{2048} $\times$ 642 $\times$ 1 \\
\hline 
\multirow{2}{*}{Global} & Pooling & - & 2048 $\times$ 1 $\times$ 1   \\
                        & Repeat & - & \textbf{2048} $\times$ 642 $\times$ 1 \\
\hline 
Neighbor & Matrix  & - & \textbf{2048} $\times$ 642 $\times$ 1   \\
(Optional) & Multiplication & - &   \\
\hline 
Output      & Concat    & - & 6147 $\times$ 642 $\times$ 1 \\
\shline
\end{tabular}}}
\end{table}

\section{More Ablation Studies}

\noindent\textbf{Person Re-id Implementation and Discussion.} 
One interesting problem remains whether our learned 3D human model can facilitate downstream tasks, such as person re-identification (re-id), which intends to match the pedestrian from different viewpoints. 
To verify this point, we conduct a preliminary experiment via dataset augmentation. 
In particular, we randomly mix up the texture from two different real identities to form one new virtual pedestrian identity class, and then render five projections of the 3D mesh from -60$^{\circ}$, -30$^{\circ}$, 0$^{\circ}$, 30$^{\circ}$, and 60$^{\circ}$ rotation.
As a result, we obtain  77,566 training images of 12,991 identities (751 real identities + 12,240 virtual identities), which enlarge about 6 times images compared with the original Market training set. 
We train the model with the common setting, such as dropout, random erasing, and horizontal flipping ~\cite{zhong2020random,sun2018beyond}. The final feature dimension is set as 2048 instead of 512 to preserve more visual clues. For a fair comparison, we try our best to tune the baseline and select the most competitive hyper-parameter to report a reliable and competitive baseline result (see Table~\ref{table:baseline}).  
 For the ResNet50-ibn baseline, we train the model with a learning rate of 0.004, batch size of 16, and dropout of 0.75. 
 Similarly, for the HRNet-w18 baseline, we train the model with a learning rate of 0.005, batch size of 16, and dropout of 0.75. On the other hand, since we add many inter-class variants via 3D generation, we train the baseline on our generated dataset (Original+3D) with a smaller dropout rate of 0.2. We also set a larger batch size for faster training. For ResNet50-ibn, we train the model on the generated data with a learning rate of 0.045, and batchsize of 192. Since the parameter of HRNet-w18 is larger than ResNet50-ibn, due to the GPU memory limitation, we set batchsize as 160 with a learning rate of 0.055. 

\begin{figure}[tbp]
\begin{center}
    \includegraphics[width=0.95\linewidth]{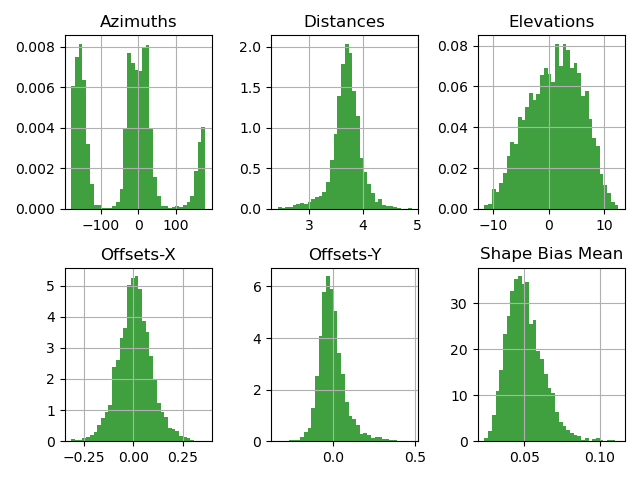}
\end{center}
\vspace{-.2in}
     \caption{ Histogram of 3D Camera Attributes $C$ on Market-HQ. Here we show the distribution of azimuths, distances, elevations, Offsets-X and Offsets-Y. Besides, we also provide the distribution of the mean shape offset $\Delta S$ over the test set. }\label{fig:Statistics}
\end{figure}

\noindent\textbf{Camera Attribute Distribution.} We observe that the camera encoder successfully captures the camera distribution in the Market-HQ test set. As shown in Figure~\ref{fig:Statistics}, most samples are $2\sim4$ unit distances from the camera, and most persons are in the center of the figure with 0 X-Offsets and 0 Y-Offsets. Most samples appear in  0$^\circ$, 180$^\circ$ or -180$^\circ$, which means that most people are facing towards or backward to the camera. It is aligned with the dataset setup since cameras are set up in front of the supermarket entrance or exit. The elevation of most samples is from -10 to 10, which is also aligned with the data collection setup, \ie, six horizontal-view cameras. Besides, we also show the distribution of mean shape offset $\Delta S$. We observe that most deformations are relatively small since we have introduced the 3D prior template. In a summary, the learned attribute statistics verify that we disentangle the camera hyper-parameter, \eg, scale changes and position offsets, from the shape encoder. 

\noindent\textbf{SMPL Initialization.}
Actually, our method is compatible with SMPL initialization, but it is worth noting that we still need to carefully consider several engineering problems. 
(1) Directly using SMPL as a general prototype? We failed. Most people in our datasets are walking with arm close to their bodies.  The model does not converge, since the rising arm is in the canonical SMPL model. 
(2) Why not use SMPL model estimation for every individual people instead of a global prototype? 
Yes. It is a great idea. We apply the state-of-the-art ROMP~\cite{ROMP} to obtain the SMPL mesh for every image as the initial shape.  However, another problem arises. The SMPL model contains too many vertexes (12,943) than our basic ellipsoid (642). It also arises the optimization problem during training. The model also does not converge.   
In the future, we would like to consider other down-sampling strategies.
(3) In the future, we may also try LBS and inverse LBS functions like~\cite{EVA3D} to conduct canonical mapping, and it may align the representation to deal with the optimization problem. However, these techniques may be beyond our work. Therefore, we leave them as future work. 

\noindent\textbf{About EMA.} 
We do not report the result with the weight moving average for a fair comparison with other methods in the main paper. EMA~\cite{tarvainen2017mean} actually can further boost our performance, and we conduct EMA for the last 100 training epochs. 
As shown in Table~\ref{table:ema}, there are significant improvements in SSIM and MaskIoU, while FID is fluctuating. Actually, it is close to our observation. No matter whether we apply EMA, we observe that the visual results are still close to the ones without EMA. EMA successfully replaces some visual changes with a more stable prediction.

\setlength{\tabcolsep}{2pt}
\begin{table}
\caption{Ablation Study of EMA on Market-HQ and ATR.
}\vspace{-.1in}
\label{table:ema}\vspace{-.15in}\footnotesize
\begin{center}\resizebox{\linewidth}{!}{
\begin{tabular}{c|c|c|c|c|c|c|c|c|c}
\hline 
\multirow{3}{*}{Methods} & \multicolumn{5}{c|}{Market-HQ} & \multicolumn{4}{c}{ATR}\\
 & MaskIoU        & SSIM            & \multirow{2}{*}{FID$_{recon}$  $\downarrow$}    & \multirow{2}{*}{FID$_{novel}$ $\downarrow$} &  \multirow{2}{*}{FID$_{90}$ $\downarrow$} & MaskIoU        & SSIM            & \multirow{2}{*}{FID$_{recon}$  $\downarrow$}    & \multirow{2}{*}{FID$_{novel}$ $\downarrow$} \\
& (\%) $\uparrow$ & (\%) $\uparrow$ &  &  &  & (\%) $\uparrow$ & (\%) $\uparrow$ &  &  \\ 
\shline
Ours & 83.4 & 66.3 & 21.5 & 46.7 & 93.3 & 81.1 & 72.6 & 35.9 & 66.8 \\ 
Ours +EMA & 87.1 & 74.0 & 20.0 & 47.3 & 94.3 & 84.3 & 80.3 & 32.6 & 65.4 \\ 
\hline
\end{tabular}}
\end{center}\vspace{-.1in}
\end{table}

\end{document}